\documentclass[runningheads]{llncs}
\usepackage[T1]{fontenc}
\usepackage{amssymb}

\usepackage[misc]{ifsym}

\newcommand{\stitle}[1]{\vspace{2mm}\noindent\textbf{#1}}
\renewcommand{\vec}[1]{\ensuremath{\mathbf{#1}}}
\usepackage{caption}
\usepackage{graphicx}
\usepackage{algorithm}
\usepackage{algorithmic}
\usepackage{amsmath}
\usepackage[shortlabels]{enumitem}
\usepackage{bm}
\usepackage{multirow}
\usepackage{tablefootnote}
\usepackage{booktabs}
\usepackage[section]{placeins}
\usepackage{fancyhdr}
\usepackage{float}
\usepackage{subfig}
\usepackage{tikz}
\usepackage{graphicx}
\usetikzlibrary{positioning}
\usepackage{tabu}
\usepackage{newfloat}
\usepackage{listings}
\usepackage{mfirstuc}

\usepackage{amssymb}
\usepackage{enumitem}
\usepackage{amssymb}
\newcommand{\ssc}[1]{\textsc{\capitalisewords{\MakeLowercase{#1}}}}
\newcommand{\model}{\textsc{DANS}}

\newcommand{\eg}{{\it e.g.}}

\newcommand{\ie}{{\it i.e.}}

\newcommand{\ent}[1]{\ensuremath{\mathtt{#1}}}
\renewcommand{\vec}[1]{\ensuremath{\mathbf{#1}}}%
\begin{document}
\title{Diversified and Adaptive Negative Sampling on Knowledge Graphs
}
%
%
\author{Ran Liu$^1$ \and
Zhongzhou Liu$^1$ \and
Xiaoli Li$^2$\and
Hao Wu$^3$\Letter \and
Yuan Fang$^{1}$\Letter}
\authorrunning{R.~Liu et al.}
%
\institute{Singapore Management University, 81 Victoria St, Singapore 188065 \\ 
\email{\{ran.liu.2020@phdcs.,zzliu.2020@phdcs.,yfang@\}smu.edu.sg} 
\and Institute for Infocomm Research, 1 Fusionopolis Way, Singapore 138632 \\ \email{xlli@i2r.a-star.edu.sg}
\and Beijing Normal University, 19 Xinwai Ave, Haidian District, Beijing 100875 \\
\email{wuhao@bnu.edu.cn}}
\maketitle              
\pagestyle{empty}
\begin{abstract}
In knowledge graph embedding, aside from positive triplets (\ie, facts in the knowledge graph), the negative triplets used for training also have a direct influence on the model performance. In reality, since knowledge graphs are sparse and incomplete, negative triplets often lack explicit labels, and thus they are often obtained from various sampling strategies (\eg, randomly replacing an entity in a positive triplet). An ideal sampled negative triplet should be informative enough to help the model train better. However, existing methods often ignore \emph{diversity} and \emph{adaptiveness} in their sampling process, which harms the informativeness of negative triplets. As such, we propose a generative adversarial approach called \textbf{D}iversified and \textbf{A}daptive \textbf{N}egative \textbf{S}ampling (\model) on knowledge graphs. \model\ is equipped with a two-way generator that generates more diverse negative triplets through two pathways, and an adaptive mechanism that produces more fine-grained examples by localizing the global generator for different entities and relations. On the one hand, the two-way generator increase the overall informativeness with more diverse negative examples; on the other hand, the adaptive mechanism increases the individual sample-wise informativeness with more fine-grained sampling. Finally, we evaluate the performance of \model\ on three benchmark knowledge graphs to demonstrate its effectiveness through quantitative and qualitative experiments.
\keywords{Knowledge graphs\and Graph representation learning\and Graph neural networks\and Negative sampling.}
\end{abstract}
\section{Introduction}
Knowledge graphs have been widely used to encode facts about the real world. Typically, each fact describes a relationship between a head and tail entity in the form of a triplet $\langle$head, relation, tail$\rangle$, and different entities across facts are interconnected to form a graph structure.
The rich facts contained in a large-scale knowledge graph can be used to enhance numerous applications that rely on real-world knowledge, such as question answering \cite{wu2016ask,li2021improving}, object detection \cite{fang2017object} and recommendation \cite{cao2019unifying}. To effectively exploit the facts for these applications, a common approach is to first perform knowledge graph embedding that converts the symbolic entities and relations to a latent vector space, 
which can then be integrated with other machine learning models.

In this paper, we focus on the problem of \emph{knowledge graph embedding}. The high-level idea is that the embedding vectors of entities and relations co-occurring in the same fact should be bounded by certain constraints due to their relatedness. For instance, consider a fact 
$\tau=\langle h=\ent{Beijing}, r=\ent{isCapitalOf}, t=\ent{China}\rangle$
and a classic method TransE \cite{bordes2013translating}. TransE maps each entity and relation in the fact to vectors, \ie, $\vec{e}_h$, $\vec{e}_r$, $\vec{e}_t$, respectively, so that they approximately satisfy the constraint $\vec{e}_h+\vec{e}_r\approx \vec{e}_t$ by minimizing the loss $\|\vec{e}_h+\vec{e}_r - \vec{e}_t\|$. On the contrary, a nonfact such as $\langle h=\ent{Beijing}, r=\ent{isCapitalOf}, t'=\ent{Russia}\rangle$ would maximize the loss $\|\vec{e}_h+\vec{e}_r - \vec{e}_{t'}\|$. Given this contrast, the factual triplets are known as \emph{positive} triplets (or examples), whereas the non-factual triplets are called \emph{negative} triplets. 
Although positive triplets are readily available, negative triplets are often obtained through random sampling.
More recent works \cite{xiong2017deeppath,cai2017kbgan} explore advanced constraints or losses \cite{bordes2013translating,trouillon2016complex} on the triplets, but the sampling strategy for negative triplets remains a crucial yet less explored problem.

Earlier negative sampling approaches resort to random sampling, \eg, by replacing the tail (or head) entity in a positive triplet with a random entity from the knowledge graph sampled in a uniform \cite{xi2021deep} or popularity-weighted manner \cite{mikolov2013distributed}.
Although random sampling is straightforward, it is often inadequate to optimize the \emph{informativeness} of negative triplets.
The informativeness refers to how much information each negative triplet could contribute to model learning.
Intuitively, a more informative negative triplet would improve the efficiency of model training and accelerate model convergence. 
For instance, the positive triplet $\tau$ given earlier, $\langle$\ent{Beijing}, \ent{isCapitalOf}, \ent{Russia}$\rangle$ is considered a more informative negative triplet than $\langle$\ent{Beijing}, \ent{isCapitalOf}, \ent{KFC}$\rangle$, as the latter can be easily identified as negative and thus helps little in refining the decision boundary. Although various scoring functions \cite{bordes2013translating,trouillon2016complex,yang2014embedding} help to judge the informativeness of negative triplets, they do not consider the \emph{diversity} and \emph{adaptiveness} of the sampling process, which are two aspects we propose to study in this work. 

On one hand, diversity helps to increase the \emph{overall} informativeness of all the negative triplets collectively.
We observe that negative triplets can be associated with both entities and relations.
For example, the tail entity of the positive triplet $\tau$ can be replaced by entities associated not only with the head entity \ent{Beijing}, such as \ent{GreatWall} and \ent{Shanghai}, but also with the relation \ent{isCaptialOf}, such as \ent{Russia} (a country with some capital city) and \ent{London} (a capital city of some country). On the other hand, adaptive sampling of negative triplets would make entity- or relation-specific adjustments to sample selection, which increases the \emph{individual} informativeness of each triplet in a finer-grained manner. For instance, selecting a tail entity for \ent{Beijing}, \ent{Tokyo} or \ent{KFC} using a global sampling model could be suboptimal given the variability among these entities. Instead, local models that condition on each entity would be able to adapt to such differences and make each triplet more informative.

In view of the above, we propose a \textbf{D}iversified and \textbf{A}daptive \textbf{N}egative \textbf{S}ampling (\model) approach for knowledge graph embedding, to improve both the overall and individual informativeness of negative triplets. Similar to previous state-of-the-art approaches such as KBGAN \cite{cai2017kbgan}, we adopt  a generative adversarial network (GAN) \cite{wang2018incorporating} for the generation of negative samples. However, there are two significant differences from previous GAN-based negative sampling on the knowledge graph. First, we design a two-way generator to produce diversified samples that are associated with both entities and relations w.r.t.~a positive triplet, which aims to increase the overall informativeness of the samples. More specifically, the generator consists of two pathways 
to produce two different kinds of negative triplets associated with a given entity and entity-relation, respectively. Second, we design an adaptive mechanism to modulate the global generator model into local models to handle the differences across entities and relations, which aims to increase the individual informativeness of the samples in a finer-grained manner. In particular, we employ a Feature-wise Linear Modulation (FiLM) layer \cite{perez2018film} that conditions the generator on a given entity or entity-relation input. 

In summary, we make the following contributions. (1) We design a two-way generator to produce diverse negative triplets, to increase the overall informativeness. (2) We employ a FiLM layer to adapt the global generator model into local models, to increase the individual informativeness of the negative triplets. (3) We conduct extensive experiments on three benchmark datasets. The results demonstrate the superiority of our proposed approach. 

\section{Background}\label{sec:relatedwork}
Negative sampling is an important issue in various machine learning tasks such as recommendation systems \cite{rendle2012bpr} and natural language processing \cite{mikolov2013distributed}.
In the context of knowledge graph embedding, negative triplets are often constructed by replacing the tail or head entity in a positive triplet with a randomly sampled entity \cite{bordes2013translating}. Unfortunately, in uniform \cite{bordes2013translating} or popularity-weighted sampling \cite{mikolov2013distributed}, the sampled entity could be completely unrelated to the head or the relation, and therefore be less informative. 

To sample more informative negative triplets, researchers have leveraged different heuristics or learning strategies. Structure-aware models \cite{ahrabian2020structure,ying2018graph} exploit the graph structures, which generally select negative examples in the neighborhood of positive examples. For example, SANS \cite{ahrabian2020structure} hypothesizes that entities that are in close proximity to each other, but do not share a direct relationship, are better candidates for negative sampling. 
In a similar spirit, PinSage \cite{liu2020aggregating} generates localized graphs via random walks to extract informative negative samples. However, these approaches have a high risk of selecting false negatives, as not explicitly related entities in close proximity could still form positive triplets due to the incompleteness of the observed graph. 

Other approaches seek to quantify the informativeness of the negative triplets through various learning strategies, including GANs \cite{cai2017kbgan,wang2018incorporating}, reinforcement learning \cite{wang2020reinforced}, and importance sampling \cite{zhang2019nscaching}. These methods provide a more explicit and systematic scoring of negative triplets which often led to better performance. However, these approaches do not consider the diversity and adaptiveness of negative sampling, which are crucial to the overall and individual informativeness of the negative triplets, respectively. 

Besides, recent studies \cite{yang2020understanding,qian2021understanding} show that the optimal negative sampling distribution should be positively but sub-linearly correlated to the positive sampling distribution. Although our proposed model shares a similar view by learning the underlying distribution of positive samples to produce negative samples, we take one step further to consider the diversity and adaptiveness of the negative samples in an adversarial setting. In particular, toward adaptiveness, we borrow the idea from Feature-wise Linear Modulation (FiLM) \cite{perez2018film}, which was first  introduced in the area of visual question answering. Its mechanism includes a learnable feature-wise affine transformation on the hidden neurons of a neural network, conditioned on an arbitrary input. In our context, we employ a FiLM layer to adapt the global generators into local models conditioned on individual input (entity or relation).

\section{Methodology} \label{sec:methodology}
In this section, we introduce the problem formulation and some preliminaries on knowledge graph embedding, followed by our proposed approach \model. 
\begin{figure*}[tbp]
    \centering
    \includegraphics[width=1.0\linewidth]{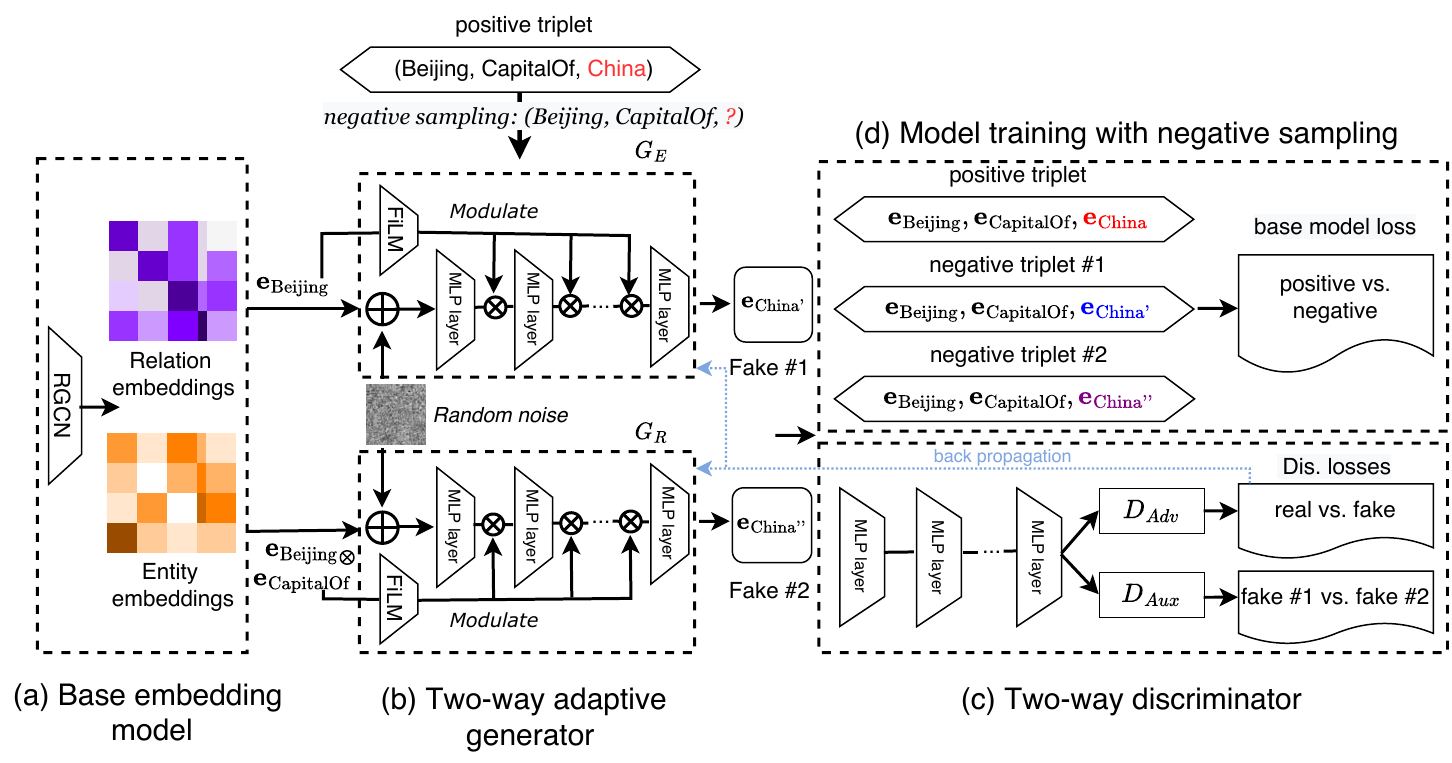}%
    \caption{Overall framework of \model. The toy example only shows how to generate fake tail entities, while generating fake head entities follows a similar process.}
    \label{fig:framework}
\end{figure*}

Before we delve into the details, we first sketch the overall framework in Figure~\ref{fig:framework}. The model consists of four main parts:  (a) A \textit{base embedding model} which learns the embeddings for entities and relations; (b) the \textit{two-way adaptive generator} which generate ``fake'' entity samples to construct negative examples; (c) the \textit{two-way discriminator} which utilize both adversarial and auxiliary losses to improve the quality of produced samples; (d) \textit{model training with negative sampling}, where we replace one entity in a positive triplet with a generated fake entity to form negative triplets, and train the base model together with the original positive triplets.

\subsection{Problem formulation and preliminaries}
A knowledge graph (KG) is defined by an entity (node) set $\mathcal{V}$, a relation set $\mathcal{R}$ and a ground-truth or positive triplet (edge) set $\mathcal{E}$. Given a triplet $\tau=\langle h,r,t\rangle$ for some $h, t\in \mathcal{V}$ and $r\in \mathcal{R}$,
a typical KG model aims to learn a scoring function 
$\mathcal{F}(\tau)$
to estimate the probability that $\tau$ is a positive triplet, \ie, $\tau$ is a fact that should appear in the ground truth set $\mathcal{E}$. 

Given the power of graph convolutional networks, in this paper, we adopt a multi-layer relational graph convolutional network (RGCN) \cite{schlichtkrull2018modeling} to serve as our base embedding model in Figure~\ref{fig:framework}(a). The base model encodes the entities in layer $l+1$ into vectors $\vec{e}_i^{l+1}\in \mathbb{R}^{d^{l+1}}$ in a latent embedding space, by aggregating their embeddings $\vec{e}_j^{l}\in \mathbb{R}^{d^{l}}$ from the previous layer $l$, as follows.
\begin{align}
    \vec{e}_i^{l+1}=\textsc{ReLU}\textstyle\left(\sum_{r \in \mathcal{R}} \sum_{j \in \mathcal{N}_i^r} \frac{1}{|\mathcal{N}_i^r|} W_r^{l} \vec{e}_j^{l}+W_0^{l} \vec{e}_i^{l}\right), 
\end{align}
where $\mathcal{N}_i^r$ is the set of neighbors of entity $i$ under relation $r$, $W_{r}^{l}$ is a trainable weight matrix for $r$, $W_0^{l}$ is an additional trainable weight matrix to capture the self-information of each entity in layer $l$, and \textsc{ReLU} is the activation function. Assuming a total of $L$ layers are stacked, the embeddings in the last layer are the output embeddings, which we simply write as $\vec{e}_i\in\mathbb{R}^d,\forall i\in \mathcal{V}$.

To optimize the parameters, a set of training triplets $\mathcal{D}_\text{tr}$ that consists of both positive and negative triplets is used.
As shown in Figure~\ref{fig:framework}(d), our objective is to sample a set of high-quality negative triplets, which, together with positive triplets, will be used to minimize the following cross-entropy loss:
\begin{align}
\label{eq:loss_base}
    \textstyle- \sum_{\tau\in \mathcal{D}_\text{tr}}y_{\tau} \log \mathcal{F}(\tau)
    +\left(1-y_\tau\right) \log \left(1-\mathcal{F}(\tau)\right)\,
\end{align}
where $y_\tau=1$ if $\tau \in \mathcal{E}$, else $y_\tau=0$. We implement $\mathcal{F}$ using three popular decoders, namely, DistMult \cite{yang2014embedding}, ComplEx \cite{trouillon2016complex}  and RotatE \cite{sun2019rotate}. We provide the DistMult function below, and leave the details of ComplEx and RotatE to Appendix A of the extended version \cite{liu2024diversifiedadaptivenegativesampling}. 
\begin{align}
    \mathcal{F}(\langle h,r,t \rangle)=\sigma(\vec{e}_h^\top \text{Diag}(\vec{e}_r)  \vec{e}_t), 
\end{align}
where $\sigma$ is the sigmoid activation, $\vec{e}_h,\vec{e}_t$ are the head, tail entity embeddings from RGCN, $\text{Diag}(\vec{e}_r) \in \mathbb{R}^{d\times d}$ is diagonal matrix whose diagonal is $\vec{e}_r$, an $r$-specific trainable vector of the decoder. Therefore, the full set of training parameters of the base model is
$\Theta=\{W_{r}^{l}:r\in\mathcal{R},l\le L\} \cup \{W_0^l:l\le L\} \cup \{\vec{e}_r:r\in\mathcal{R}\}.$

\subsection{Adaptive two-way generator}
A common way to obtain a negative triplet is to replace the tail (or head) entity in a positive triplet by a randomly sampled entity. Beyond simple random sampling, generative adversarial nets (GAN) \cite{goodfellow2014generative} such as KBGAN \cite{cai2017kbgan}, IGAN \cite{wang2018incorporating}, HeGAN \cite{hu2019adversarial} and GNDN \cite{zeng2020learning}, which learn the underlying sample distributions, have been shown to be effective in negative sampling on KG or other graph structures. 

Formally, given a positive triplet $\langle h,r,t\rangle$, a generator $G$ aims to produce a ``fake'' tail entity $t'$ to replace the real tail $t$, resulting in a negative triplet  $\langle h,r,t' \rangle$. 
More precisely, $G$ is a function that maps a noise $\epsilon$ (typically sampled from a prior distribution) to a vector $\vec{e}_{t'}$ in the entity embedding space. Although we follow a similar process, distinct from existing GAN-based approaches, we propose an adaptive two-way generator, as shown in Figure~\ref{fig:framework}(b). It not only diversifies the generation of fake entities, but also localizes the global generator model to adapt to fine-grained differences across entities. 

\stitle{Diversity.} Classical GANs generate fake samples through a single pathway and assume a fixed prior distribution, which limits the diversity of fake entity generation. Particularly, in the context of KG, we can generate a fake tail entity associated with either the head entity only, or the relation as well. This improves the diversity of resulting negative triplets and increases the overall informativeness. Hence, we propose a two-way generator that consists of two pathways, namely $G_E$ and $G_R$, to generate negative triplets associated with a given entity and entity-relation, respectively. Furthermore, 
having personalized priors for each entity or relation would further enhance the diversification. Specifically, to replace the tail entity in a positive triplet $\langle h,r,t \rangle$ (the same process would also apply to replacing the head entity $h$), we generate fake tail entity embeddings $\vec{e}_{t'}$ and $\vec{e}_{t''}$ from the two pathways, as follows.
\begin{align}
\label{eq:fake1}
    \vec{e}_{t'} &= G_E(\epsilon;\Theta_{G_E}), \ \text{s.t. } \epsilon \sim N(\vec{e}_h,\sigma^2I),\\
    \label{eq:fake2}
     \vec{e}_{t''} &= G_R(\epsilon;\Theta_{G_R}), \ \text{s.t. } \epsilon \sim N(\vec{e}_h \otimes \vec{e}_r,\sigma^2I),
\end{align}
where each pathway has its own parameters, \ie, $G_E$ parameterized by $\Theta_{G_E}$ and $G_R$ parameterized by $\Theta_{G_R}$. The noise vector $\epsilon$ that feeds into each pathway is sampled from a personalized multivariate Gaussian distribution for each entity/relation, $N(\vec{e}_h,\sigma^2I)$ or $N(\vec{e}_h \otimes \vec{e}_r,\sigma^2I)$ depending on the pathway. $N$ represents the prior Gaussian distribution for sampling the input to the generator $\epsilon$, $\sigma$ is a hyper-parameter controlling the covariance of the multivariate Gaussian, $I$ is the identity matrix, and $\otimes$ stands for element-wise multiplication. Intuitively, as the prior Gaussian distributions in Eqs.~\eqref{eq:fake1} and \eqref{eq:fake2} are centered on different embeddings,
$\vec{e}_h$ or $\vec{e}_h \otimes \vec{e}_r$, it helps to diversify the generated samples from different pathways.

Each pathway is implemented as a multi-layer perceptron (MLP). Taking $G_E$ as an example, its MLP is parameterized by $\Theta_{G_E}$ which consists of the weights and biases in each layer. Let $\vec{x}_{G_E}^{m+1}$ denote the activations of the $m$-th MLP layer, where the activations of the last MLP layer are simply the output embedding of $G_E$. The architecture of $G_R$ mirrors that of $G_E$.

\stitle{Adaptiveness.} 
While more diverse samples help increase the overall informativeness, it is also important to improve the informativeness of individual samples. On the one hand, all input entities or relations sharing a global generator model are unable to fully adapt to fine-grained differences across entities or relations. On the other hand, training one model for each entity or relation can cause severe overfitting and incur large overheads. To address the dilemma, we still train a global generator model, but allow the global model to be modulated through a Feature-wise Linear Modulation (FiLM) layer conditioned on each input entity or relation, which essentially adapts the shared global model into local models. Thus, in addition to the global model parameters, the adaptive mechanism only needs to learn the parameters of the FiLM layer, instead of one set of model parameters for each entity or relation.

Consider the pathway $G_E$ to generate a fake tail entity for a head entity $h$. We adapt the global model $G_E$ to suit the head entity $h$, by modulating the activations in each hidden layer of $G_E$:
\begin{align}
    \tilde{\vec{x}}_{G_E}^{m} = \vec{x}_{G_E}^{m} \otimes \alpha_h^m + \beta_h^m,
\end{align}
where $\alpha_h^m$ and $\beta_h^m$ are vectors conditioned on the head entity $h$ and have the same dimension as the $m$-th layer of $G_E$. They are used to scale and shift the activations $\vec{x}_{G_E}^{m}$ of the $m$-th layer of $G_E$. That is, the global $G_E$ is adapted into a local model conditioned on $h$.
More specifically, $\alpha_h^m$ and $\beta_h^m$ are
output of the FiLM layer $F_E$ applied to the $m$-th layer of $G_E$, as follows.
\begin{align}
    \alpha_h^m = F_E(\vec{e}_h;\Theta_{F_E,\alpha}^m),\\
    \beta_h^m = F_E(\vec{e}_h;\Theta_{F_E,\beta}^m).
\end{align}
Note that the head entity embedding $\vec{e}_h$ is the input to $F_E$, making the output adaptive to and conditioned on $h$.
$F_E$ can be implemented as a MLP, parameterized by $\Theta_{F_E,\alpha}^m$ and $\Theta_{F_E,\beta}^m$ in the $m$-th layer of $G_E$.  Similarly, the second pathway $G_R$ can be modulated by a FiLM layer $F_R$, whose input is $\vec{e}_h \otimes \vec{e}_r$, to generate a fake tail entity for a head entity $h$ and relation $r$. $F_R$ is parameterized by $\Theta_{F_R,\alpha}^m$ and $\Theta_{F_R,\beta}^m$ in the $m$-th layer of $G_R$, to output $\alpha_{h,r}^m$ and $\beta_{h,r}^m$ to scale and shift the activations in $G_R$. 

To sum up, the trainable parameters in the adaptive two-way generator, $\Theta_G$, include the weights in the two global pathways and the FiLM layer weights for each layer in each pathway. For a total of $M$ hidden layers in the global pathways, we have $\Theta_G = \{\Theta_{G_E}, \Theta_{G_R}\} \cup  \{\Theta_{F_E,\alpha}^m,\Theta_{F_E,\beta}^m,\Theta_{F_R,\alpha}^m,\Theta_{F_R,\beta}^m:  m \le M\}$.
\subsection{Two-way discriminator}
As in a standard GAN, a discriminator is needed to help the generator produce high-quality fake entities that mimic real entities. Specifically, the discriminator and the generator compete with each other in a minimax game, in which the generator aims to fool the discriminator by producing realistic looking entities, while the discriminator aims to beat the generator by distinguishing the real and fake entities. In our case, given the two-way generator, we further equip the discriminator with the ability to distinguish the fake entities generated by the two pathways, which can further differentiate and diversify the two pathways. 



Concretely, as shown in Figure~\ref{fig:framework}(c), the discriminator also has two pathways: $D_{\text{Adv}}$, an \textit{adversarial} pathway to distinguish fake and real entities, and $D_{\text{Aux}}$, an \textit{auxiliary} pathway to distinguish fake entities generated by $G_E$ and $G_R$. 
Taking the generation of tail entities as an example, given the real tail entity $t$ in a positive triplet, as well as the fake entities $t'$ generated by $G_E$ and $t''$ generated by $G_R$, $D_{\text{Adv}}$ tries to distinguish $t$ from $t'$ and $t''$, while $D_{\text{Aux}}$ tries to distinguish $t'$ from $t''$. In other words, each of them involves a binary classification:
\begin{align}
    \hat{y}_{\text{Adv},i} &= D_{\text{Adv}}(\tilde{\vec{e}}_i;\Theta_{D_{\text{Adv}}}),\\
    \hat{y}_{\text{Aux},i} &= D_{\text{Aux}}(\tilde{\vec{e}}_i;\Theta_{D_{\text{Aux}}}), \text{ \ s.t. } i \neq t,
\end{align}
where $D_{\text{Adv}}$ and $D_{\text{Aux}}$ are implemented as a fully connected layer, and $\tilde{\vec{e}}_i=\ssc{MLP}(\vec{e}_i; \Theta_{D_{S}})$ is a shared hidden representation computed from the embedding $\vec{e}_i$ of a real or fake entity $i$. 
The shared hidden representation allows both $D_{\text{Adv}}$ and $D_{\text{Aux}}$ to benefit from each other during training \cite{odena2016semi}, as they collectively try to distinguish three different classes of samples ($t$, $t'$, $t''$).

Note that $\hat{y}_{\text{Adv},i}$ (or $\hat{y}_{\text{Aux},i}$) is the predicted value of the ground-truth label $y_{\text{Adv},i}$ (or $y_{\text{Aux},i}$), such that $y_{\text{Adv},i}=1$ if $i$ is a real entity, else $y_{\text{Adv},i}=0$. Furthermore, for a fake entity $i$, we define $y_{\text{Aux},i}=1$ if $\vec{e}_i$ is generated via Eq.~\eqref{eq:fake1}, or 0 if generated via Eq.~\eqref{eq:fake2}. Subsequently, we employ a cross-entropy loss on the two discriminator pathways: 
\begin{flalign}
\mathcal{L}_{\text{Adv}}(\hat{y}_{\text{Adv},i}, y_{\text{Adv},i})  =  - y_{\text{Adv},i}\log \hat{y}_{\text{Adv},i} - (1-y_{\text{Adv},i}) \log (1-\hat{y}_{\text{Adv},i}),\\
\mathcal{L}_{\text{Aux}}(\hat{y}_{\text{Aux},i}, y_{\text{Aux},i})=  - y_{\text{Aux},i}\log \hat{y}_{\text{Aux},i} - (1-y_{\text{Aux},i}) \log (1-\hat{y}_{\text{Aux},i}), 
\end{flalign}
In summary, the set of trainable parameters of the two-way discriminator pathway includes the shared parameters and the weights of each classifier, \ie, 
$\Theta_D=\{\Theta_{D_{S}},\Theta_{D_{\text{Adv}}}, \Theta_{D_{\text{Aux}}}\}.$

\subsection{Adversarial training}
Lastly, we train the generator, discriminator, and base embedding model jointly.
On the one hand, the generator aims to fool the adversarial pathway of the discriminator, making $D_{\text{Adv}}$ harder to distinguish real and fake entities, as below.
\begin{align}
\label{eq:g_step}
   \textstyle \arg\max_{\Theta_G}  & \ \mathbb{E}_{t}{\mathcal{L}_{\text{Adv}}(\hat{y}_{\text{Adv},t}, 1)} 
    + \mathbb{E}_{t'}{\mathcal{L}_{\text{Adv}}(\hat{y}_{\text{Adv},t'}, 0)} +  \mathbb{E}_{t''}{\mathcal{L}_{\text{Adv}}(\hat{y}_{\text{Adv},t''}, 0)}\nonumber\\
   & + \lambda\textstyle\sum_{m,h}(\|\alpha_h^m-\textbf{1}\|^2 + \|\beta_h^m\|^2 ),
\end{align}
where $t$ is a real tail entity, and $t',t''$ are fake tail entities from $G_E$ and $G_R$, respectively (again, we only illustrate the case where the tail entity in a positive triplet is replaced).   
The last term in Eq.~\eqref{eq:g_step} is a regularization term on the scaling and shifting factors to prevent overfitting   \cite{oreshkin2018tadam}, and $\lambda$ is a hyper-parameter to control the strength of regularization.
On the other hand, the goal of the discriminator is to overcome the generators by distinguishing fake and real entities, as well as fake entities from different generator pathways, as follows.
\begin{align}
\label{eq:d_step}
    \textstyle\arg\min_{\Theta_{D}} & \ \mathbb{E}_{i}{\mathcal{L}_{\text{Adv}}(\hat{y}_{\text{Adv},i}, y_{\text{Adv},i})}
    + \mathbb{E}_{i\neq t}{\mathcal{L}_{\text{Aux}}(\hat{y}_{\text{Aux},i}, y_{\text{Aux},i})},
\end{align}
where $i$ can be either real or fake entity in the first term, but $i\ne t$ can only be a fake entity in the second term.

Following a typical adversarial training scheme in negative sampling on knowledge graphs in KBGAN~\cite{cai2017kbgan}, we alternate the model updating among the three parties, as follows. First, we train the generator by updating the generator parameters $\Theta_{G}$ with Eq.~\eqref{eq:g_step}, while freezing the discriminator parameters $\Theta_{D}$ and the base model parameters $\Theta$. Next, we update $\Theta_{D}$ with Eq.~\eqref{eq:d_step}, while freezing $\Theta_{G},\Theta$. Finally, we update $\Theta$ by minimizing the loss on the positive and negative triples in Eq.~\eqref{eq:loss_base}, while freezing the other two parameter sets. We repeat the three steps until the convergence of all parties are achieved.

\section{Experiments}\label{sec:expt}
We perform empirical evaluation on three benchmark knowledge graphs. We first compare the empirical performance of the proposed model \model\footnote{https://github.com/liuran998/DANS} with state-of-the-art baselines. In addition, we seek to address a number of research questions (RQs). 
\textbf{RQ1}: Does the two-way design in the generator improve model performance? \textbf{RQ2}: Does the adaptive FiLM layer in the generator improve model performance? \textbf{RQ3}: What is the impact of the number of negative triplets and adaptive regularization, respectively? \textbf{RQ4}: Can we observe the diversity of generated triplets?

\subsection{Experimental Design}
\begin{table}[t]
    \centering
    \scriptsize
    \addtolength{\tabcolsep}{1mm}
    \caption{Statistics of datasets.}
    \label{table:1}
    \begin{tabular}{crrrrrr}
    \toprule
    & Entities & Relations & Train & Validation & Test & Total \\
    \midrule
    WN18RR    &  40,943 & 11 &  86,835 & 3,034   & 3,134 & 93,003 \\
    NELL-995   &  63,916  & 198 &  137,465 &  5000 & 5000 & 147,465\\
    UMLS  &  135  & 46 &  5,216 &   652  & 661 & 6,529\\
    \bottomrule
    \end{tabular}
\end{table}
\stitle{Datasets.} Three benchmark knowledge graphs are used for our experiment. (1) \textbf{WN18RR} \cite{bordes2013translating}, a harder variant of WN18 \cite{dettmers2018conve}, which is derived from WordNet consisting of hyponym and hypernym relations between words. Compared to WN18, WN18RR removes inverse relations to minimize leakage from training. (2) \textbf{NELL-995} \cite{dutta2025replacing} is a subset of the web-based facts collected by the 995th iteration of the Nell system \cite{carlson2010toward} which contains a large pool of entity types and only the top 200 relations are retained. (3) \textbf{UMLS} \cite{dettmers2018conve} is a specialized knowledge base containing medical entities and their semantic relationships. The entities are biomedical concepts (e.g., disease, antibiotic), and the relations include interactions such as \ent{treats} and \ent{diagnosis}. Table~\ref{table:1} gives a summary of the datasets.

\stitle{Task and evaluation.}
We employ the standard knowledge graph completion task \cite{nathani2019learning,cao2019unifying,ji2016knowledge,chen2020knowledge}. Specifically, for each positive test triplet, we construct a list of candidate triplets that also include negative triplets, which are obtained by replacing either the head or tail of the positive triplet with every other entity in the dataset. To avoid false negatives, we follow previous work \cite{bordes2013translating} by adopting their ``filtered setting''.
We then rank the candidate triplets based on the scoring function. For evaluation, we adopt several standard ranking metrics including Mean Reciprocal Ranking (MRR), Hit ratio at 1 (H@1) and Normalized discounted cumulative gain at 5 (NDCG@5) \cite{sun2020we}. Details of these ranking metrics can be found in Appendix B of the extended version \cite{liu2024diversifiedadaptivenegativesampling}.

\stitle{Baselines.} We compare with baselines in two distinct categories: 

(1) \emph{Negative samplers} with the same RGCN  backbone \cite{schlichtkrull2018modeling} and decoders. In other words, they are flexible ``plug-ins" that only replace the sampling strategy for a fair comparison to our method \model.  
They include \textbf{Rand}, which replaces the head or tail entity with a uniformly sampled random entity; 
\textbf{Pop} \cite{mikolov2013distributed}: a variant of Rand that substitutes uniform sampling with popularity-weighted sampling; \textbf{Self-adv} \cite{sun2020we}: a self-adversarial negative sampling methodology; \textbf{MCNS} \cite{yang2020understanding}: a model which derives negative samples from a distribution that is positively but sub-linearly correlated with the positive distribution.

(2) \emph{Other state-of-the-art baselines} for knowledge graph embedding which may employ a variety of different backbones, heuristics and techniques that diverge from \model, for a comprehensive comparison. They include \textbf{SANS-RW} \cite{ahrabian2020structure}: a structure-aware model that selects negative samples at close proximity from positive nodes via random walks on the graph; \textbf{NSCaching} \cite{zhang2019nscaching}: a model that employs importance sampling to sample more informative negative triplets; \textbf{KBGAN} \cite{cai2017kbgan}: a GAN-based model that learns to generate informative negative triplets; \textbf{CAKE} \cite{niu2022cake}: a framework which leverages extra information such as entity types to from factual triplets to sample negative triplets; \textbf{SMiLE} \cite{peng2022smile}: a framework which employs specific contextual information influenced by entity types to sample negative triplets. 

\stitle{Parameter settings.} Our model \model\ and other negative samplers (Random, Pop, Self-Adv and MCNS) employ RGCN \cite{schlichtkrull2018modeling} as the backbone, which follows JinheonBaek's pytorch implementation. RGCN is first pre-trained for 15000 epochs, and our base embedding model is then initialized using the pre-trained weights. We train the model for 5000 epochs, using a learning rate of 0.001 and a mini-batch size of 1000 for UMLS, WN18RR and NELL-995. In each mini-batch, the generator and discriminator epochs are set to 5 and 1, respectively, and their learning rates are set to 1e-3 and 1e-4, respectively. The regularization coefficient $\lambda$ for the FiLM layer in Eq. (13) is set to  1e-4 for all three datasets as it is the most optimal among candidate set \{1e-2, 1e-3, 1e-4, 1e-5, 1e-6\}. 

Furthermore, we generate $N_s=20$ negative triplets for each positive triplet, out of which the first ten negative triplets are equally split between the two generator pathways, while the remaining ten negative triplets are obtained via uniform random sampling to further increase the diversity. In all cases, either the head or tail of the positive triplets are randomly replaced with negative entities, but not both. We set the output embedding dimension $d$ to 100 for all methods, except SANS-RW, where $d$ is set to the recommended 1,000 to achieve optimal performance. RGCN, RGCN-P, RGCN-Adv and RGCN-MCNS follow the same implementation and settings per the backbone of \model. 

In addition, the hyper-parameters related to negative sampling via Metropolis-Hastings in RGCN-MCNS follow the original paper's link prediction experiments \cite{yang2020understanding}. To reduce the variance resulting from parameter initialization, the experimental results are calculated from an average of five runs with different seeds in all methods. Furthermore, every method is standardized to use the triplet loss in Eqs.(2). Other baseline settings have also been tuned according to the recommendations of the literature. Additional details can be found Appendix C of the extended version \cite{liu2024diversifiedadaptivenegativesampling}.

\subsection{Results and Analysis}

Table~\ref{table:2} reports quantitative comparison against the first category of baselines involving different negative samplers under the same backbone and decoder. Overall, our model $\model$ consistently leads to better performance for DistMult, RotatE and ComplEx decoders. This shows the robustness of our approach across various decoders.
In general, \model\ performs better than Rand and its variant Pop, showing that it is important to account for the informativeness of negative triples which are missing in random and popularity-weighted sampling. Since Self-Adv accounts for the informativeness by giving more weight to higher quality triplets, it generally outperforms Rand and Pop. It still lags behind \model\ in most cases as it ignores the concepts of diversity and adaptiveness.  The  variant MCNS shows better performance than Rand and its variant Pop but loses to \model\, as MCNS was originally designed for homogeneous graphs.

\begin{table}[t!]
\centering
\caption{Performance comparison with other negative sampling methods, which are plugged into the same backbone (RGCN) and decoders (DistMult, RotatE or ComplEx). The best results are in bold, and the runner-ups are underlined. 
}
\label{table:2} 
\hspace{-12mm}
\scriptsize
\begin{tabu}{@{}c|ccccccccc@{}} 
\toprule
Sampling & \multicolumn{3}{c}{WN18RR} & \multicolumn{3}{c}{NELL-995} & 
\multicolumn{3}{c}{UMLS} \\
\cmidrule(lr){2-4} \cmidrule(lr){5-7} \cmidrule(lr){8-10}
method & MRR & H@1 & NDCG@5 & MRR & H@1 & NDCG@5  & MRR & H@1 & NDCG@5   \\
  \midrule
  & \multicolumn{9}{c}{\small{DistMult}} \\
  \midrule
Rand  &.372$_{\pm.002}$ & \underline{.343}$_{\pm.003}$ & .369$_{\pm.005}$ &   .218$_{\pm.001}$ & .146$_{\pm.002}$ &  .219$_{\pm.002}$ &  .696$_{\pm.010}$ & .607$_{\pm.082}$ &  .693$_{\pm.007}$  \\
Pop   &    .374$_{\pm.002}$ & .342$_{\pm.002}$ & \underline{.376}$_{\pm.006}$ & .216$_{\pm.001}$ & .142$_{\pm.002}$ & .216$_{\pm.003}$ &   .680$_{\pm.009}$ & .589$_{\pm.012}$ & .692$_{\pm.005}$ \\ Self-adv  &    .370$_{\pm.007}$ & .332$_{\pm.010}$ & .373$_{\pm.006}$ & \textbf{.238}$_{\pm.003}$ & \underline{.156}$_{\pm.003}$ & \textbf{.241}$_{\pm.003}$ &   \underline{.717}$_{\pm.009}$ & \underline{.624}$_{\pm.015}$ & \textbf{.733}$_{\pm.008}$ \\
MCNS   & \underline{.376}$_{\pm.004}$ & .340$_{\pm.005}$ & .374$_{\pm.006}$ & .226$_{\pm.002}$   & .144$_{\pm.002}$ & .221$_{\pm.003}$ & .700$_{\pm.002}$   &.606$_{\pm.008}$  & .717$_{\pm.002}$ \\
DANS & \textbf{.381}$_{\pm.006}$ & \textbf{.352}$_{\pm.007}$ & \textbf{.386}$_{\pm.008}$ &   \underline{.227}$_{\pm.004}$& \textbf{.162}$_{\pm.007}$ &  \underline{.220}$_{\pm.009}$ &  \textbf{.724}$_{\pm.008}$ & \textbf{.641}$_{\pm.009}$ & \underline{.725}$_{\pm.008}$\\
  \midrule
  & \multicolumn{9}{c}{\small{RotatE}} \\
  \midrule
 Rand  &.234$_{\pm.009}$ & .110$_{\pm.003}$ & .260$_{\pm.008}$ &   .182$_{\pm.003}$ & .093$_{\pm.003}$ &  \.189$_{\pm.003}$ &  .817$_{\pm.015}$ & \underline{.683}$_{\pm.021}$ &  .855$_{\pm.013}$  \\
 Pop   &    .235$_{\pm.007}$ & .095$_{\pm.003}$ & .268$_{\pm.007}$ & .181$_{\pm.002}$ & \underline{.131}$_{\pm.002}$ & \underline{.200}$_{\pm.003}$ &   .800$_{\pm.005}$ & .673$_{\pm.010}$ & .839$_{\pm.004}$ \\
  Self-adv  &    .202$_{\pm.007}$ & .058$_{\pm.010}$ & .235$_{\pm.006}$ & .186$_{\pm.002}$ & .096$_{\pm.003}$ & .194$_{\pm.002}$ &   .809$_{\pm.007}$ & .677$_{\pm.007}$ & .848$_{\pm.007}$ \\ 
  MCNS   & \underline{.242}$_{\pm.009}$ & \underline{.132}$_{\pm.004}$ & \textbf{.288}$_{\pm.006}$ & \underline{.194}$_{\pm.003}$ & .122$_{\pm.004}$ & \underline{.200}$_{\pm.004}$ & \underline{.822}$_{\pm.005}$   & .682$_{\pm.006}$ & \textbf{.884}$_{\pm.006}$ \\
 DANS & \textbf{.249}$_{\pm.002}$ & \textbf{.154}$_{\pm.001}$ & \underline{.274}$_{\pm.003}$ &  \textbf{.195}$_{\pm.010}$ & \textbf{.135}$_{\pm.011}$ & \textbf{.208}$_{\pm.010}$ &  \textbf{.833}$_{\pm.004}$ & \textbf{.716}$_{\pm.006}$ &\underline{.866}$_{\pm.005}$\\
  \midrule
  & \multicolumn{9}{c}{\small{ComplEx}} \\
  \midrule
 Rand  &.386$_{\pm.007}$ & \underline{.346}$_{\pm.005}$ & .390$_{\pm.006}$ &   .245$_{\pm.004}$ & .172$_{\pm.003}$ &  \.251$_{\pm.006}$ &  .898$_{\pm.008}$ & .822$_{\pm.017}$ &  .920$_{\pm.015}$  \\ 
    Pop   &    .389$_{\pm.011}$ & .341$_{\pm.007}$ & .387$_{\pm.012}$ & .241$_{\pm.005}$ & .179$_{\pm.006}$ & .245$_{\pm.004}$ &   .840$_{\pm.009}$ & .747$_{\pm.009}$ & .865$_{\pm.008}$ \\
    Self-adv  &    .375$_{\pm.006}$ & .329$_{\pm.011}$ & .382$_{\pm.013}$ & \underline{.250}$_{\pm.005}$ & \underline{.181}$_{\pm.007}$ & \textbf{.277}$_{\pm.008}$ &   \underline{.908}$_{\pm.009}$ & \underline{.844}$_{\pm.006}$ & \underline{.925}$_{\pm.010}$ \\ 
    MCNS   & \underline{.392}$_{\pm.008}$ & .343$_{\pm.007}$ & \textbf{.394}$_{\pm.008}$ & .248$_{\pm.007}$ & .177$_{\pm.004}$ & \underline{.264}$_{\pm.009}$ & .879$_{\pm.007}$   & .835$_{\pm.005}$ & .892$_{\pm.011}$ \\
  DANS & \textbf{.404}$_{\pm.005}$ & \textbf{.347}$_{\pm.004}$ & \underline{.392}$_{\pm.009}$ &  \textbf{.257}$_{\pm.006}$ & \textbf{.186}$_{\pm.010}$ & .255$_{\pm.008}$ &  \textbf{.920}$_{\pm.007}$ & \textbf{.857}$_{\pm.011}$ &\textbf{.927}$_{\pm.008}$\\
\bottomrule
\end{tabu}
\hspace{-12mm}
\end{table}

Next, Table~\ref{table:3} compares \model\ with the second category of baselines. Negative sampling in SANS-RW is not relation-aware and thus performs poorly on datasets with more variety of relations, namely, NELL-995 and UMLS. In addition, KBGAN fell short for the two bigger datasets WN18RR and NELL-995 as it ignores graph structure in the sampling process. Furthermore, its adversarial training process potentially suffers from instability and degeneracy. On the other hand, NSCaching employs a more streamlined importance sampling approach, contributing to its competitive performance despite not considering graph structure for negative sampling. As CAKE and SMiLE leverage on extra side information such as entity types to enhance its performances, their experimental results deteriorate as such information are not available in standard knowledge graph completion benchmarks in this paper. 

\begin{table}[t!]
\centering
\caption{Performance comparison with baselines (all using the DistMult decoder). See Table~\ref{table:2} caption for entry styles. 
}
\label{table:3} 
\hspace{-12mm}
\scriptsize
\centering
    \begin{tabular}{@{}c|*{9}{c}@{}}
      \toprule
      & \multicolumn{3}{c}{WN18RR} & \multicolumn{3}{c}{NELL-995} & 
        \multicolumn{3}{c}{UMLS} \\
      \cmidrule(lr){2-4} \cmidrule(lr){5-7} \cmidrule(lr){8-10}
     Model & MRR & H@1 & NDCG@5 & MRR & H@1 & NDCG@5  & MRR & H@1 & NDCG@5  \\
      \midrule
     SANS-RW 
     & .349$_{\pm.010}$ & .340$_{\pm.013}$ &  .334$_{\pm.010}$ & .135$_{\pm.006}$ & .109$_{\pm.008}$ & .110$_{\pm.008}$ &   .510$_{\pm.008}$ & .369$_{\pm.009}$ & .478$_{\pm.003}$ \\
    
     NSCaching 
     & \underline{.374}$_{\pm.002}$ & .337$_{\pm.003}$ & \underline{.374}$_{\pm.002}$ &   .177$_{\pm.004}$ & \underline{.150}$_{\pm.003}$ & .140$_{\pm.002}$ &  .625$_{\pm.004}$ & .508$_{\pm.021}$ & .607$_{\pm.004}$ \\

     KBGAN 
     & .172$_{\pm.004}$ & .070$_{\pm.006}$ & .155$_{\pm.002}$ &  .170$_{\pm.002}$ & .077$_{\pm.004}$ & \underline{.195}$_{\pm.009}$ &  \underline{.680}$_{\pm.005}$ & \underline{.556}$_{\pm.023}$ & \underline{.654}$_{\pm.004}$\\
    
    CAKE 
     & .353$_{\pm.007}$ & \underline{.345}$_{\pm.005}$ & .351$_{\pm.008}$ &  \underline{.204}$_{\pm.006}$ & .130$_{\pm.007}$ & .175$_{\pm.012}$ &  .441$_{\pm.013}$ & .365$_{\pm.008}$ & .383$_{\pm.010}$
     \\
    
       SMiLE 
     & .315$_{\pm.006}$ & .291$_{\pm.007}$ & .294$_{\pm.012}$ &  .131$_{\pm.004}$ & .127$_{\pm.005}$ & .105$_{\pm.008}$ &  .414$_{\pm.015}$ & .345$_{\pm.007}$ & .372$_{\pm.013}$
     \\
     \model  & \textbf{.381}$_{\pm.006}$ & \textbf{.352}$_{\pm.007}$ & \textbf{.386}$_{\pm.008}$&  \textbf{.227}$_{\pm.004}$ & \textbf{.162}$_{\pm.007}$ & \textbf{.220}$_{\pm.009}$ &   \textbf{.724}$_{\pm.008}$ & \textbf{.641}$_{\pm.009}$ & \textbf{.725}$_{\pm.008}$ \\
      \bottomrule
    \end{tabular}
    \hspace{-12mm}
\end{table}

Overall, \model\ has obtained favourable performance, showing the importance of diversity and adaptiveness during negative sampling. We will conduct further ablation study in the next part to examine the contribution from each aspect. Finally, we have included the experimental results for dataset FB15k-237 which show favorable performance on ComplEx decoder in Appendix D of the extended version \cite{liu2024diversifiedadaptivenegativesampling}.

\subsection{Additional research questions}
\begin{figure}[tbp]
\centering
    \hspace{-20mm}    \subfloat[\centering Ablation study]{{\includegraphics[width=4.6cm]{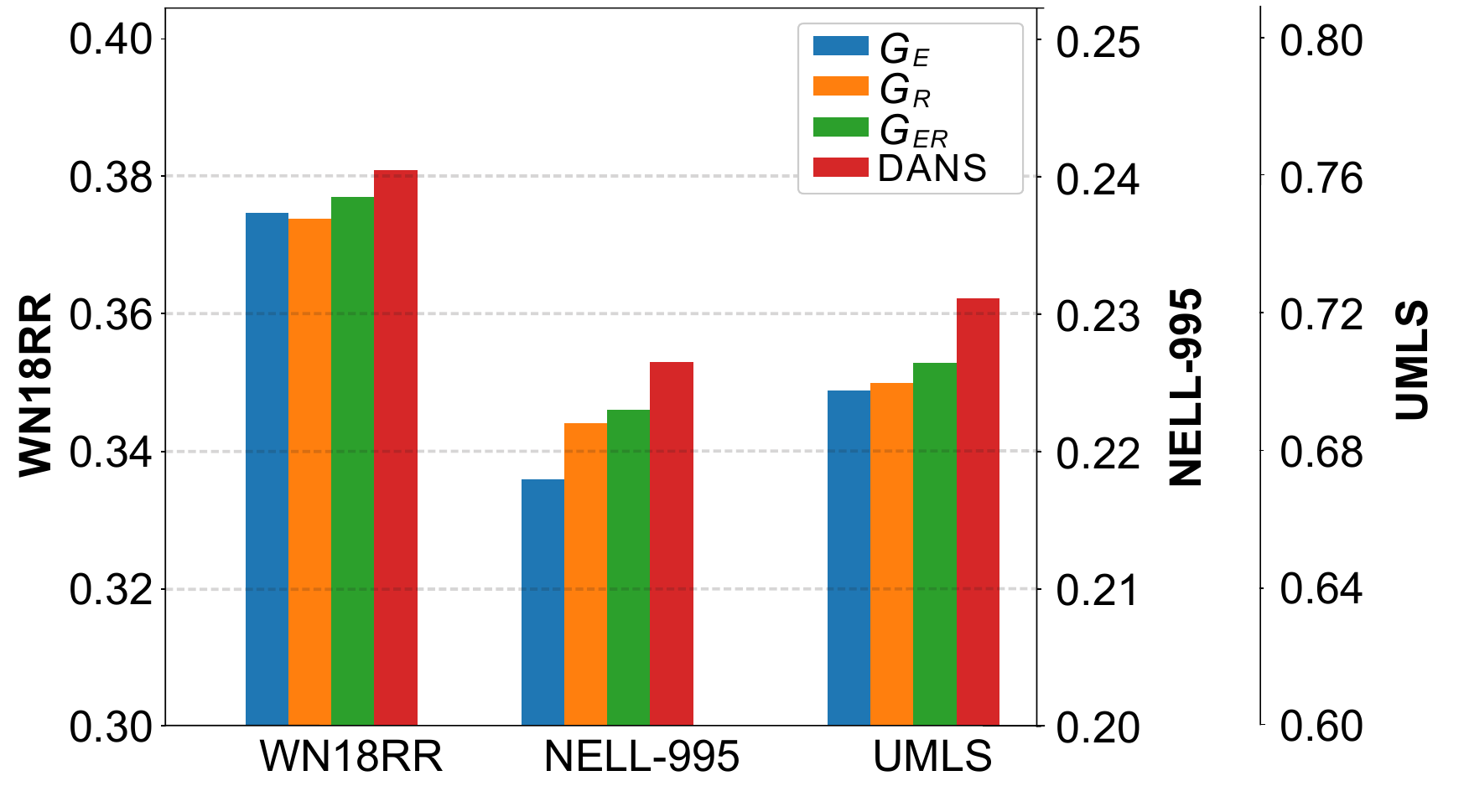} }}
    \hspace{-2mm}
    \subfloat[\centering No. of negative triplets $N_s$]{{\includegraphics[width=4.6cm]{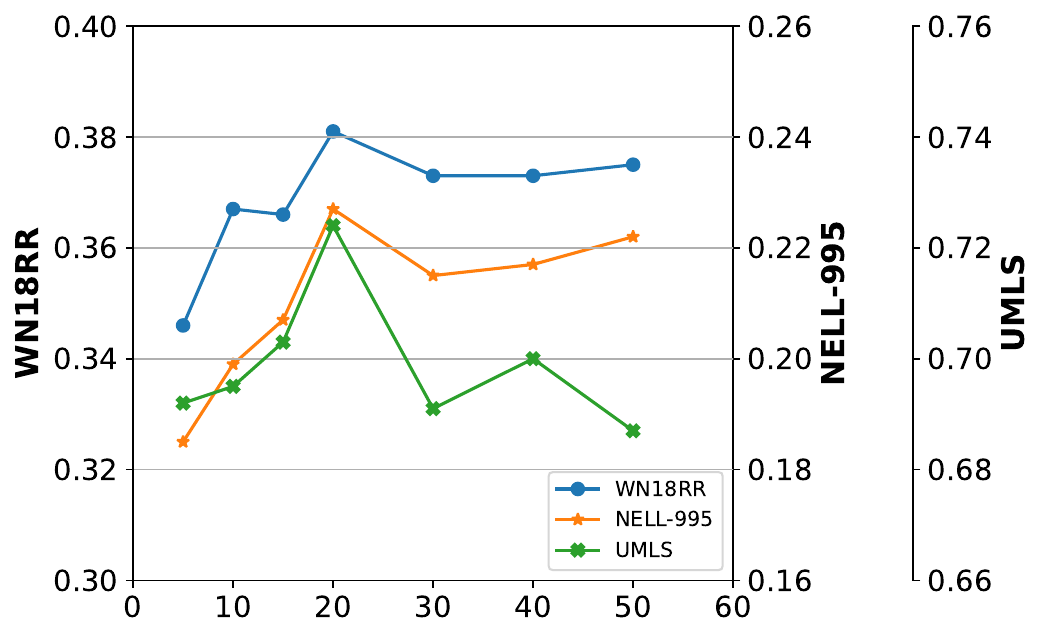} }}
    \hspace{-2mm}
    \subfloat[\centering Extent of regularization $\lambda$]{{\includegraphics[width=4.6cm]{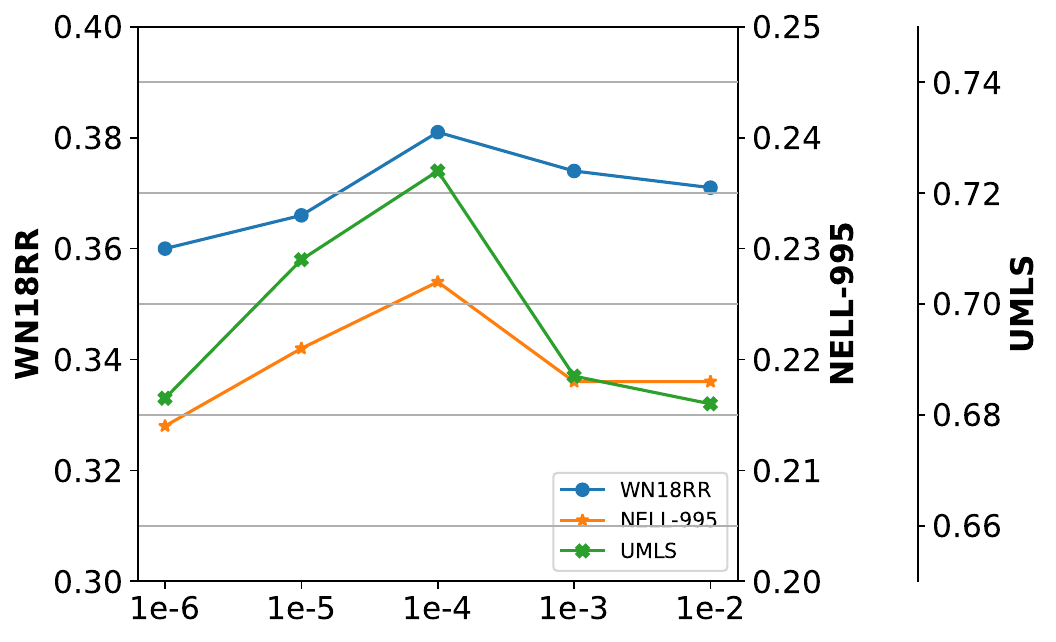} }}
    \hspace{-20mm}
    \caption{Investigation of research questions. Each dataset uses its own $y$-axis, and the metric used on the $y$-axes is MRR. (a) Ablation study on the contribution of individual generator pathways and adaptive FiLM layer. (b) Study of parameter sensitivity on the number of negative triplets $N_s$ and (c) extent of adaptive regularization $\lambda$. }\label{figure2}
\end{figure}
In this part, we seek to investigate RQ1--RQ4 listed at the beginning of this section. All experiments in this part are conducted using the DistMult function as the decoder.

\stitle{Ablation study (RQ1, RQ2).}
We investigate the contribution from major design choices through an ablation study. As depicted in Figure~\ref{figure2}(a), we compare \model\ with the following variants, all of which \emph{do not employ the FiLM layer}. (1) $G_E$: Only the pathway $G_E$ in the generator; (2) $G_R$: Only the pathway $G_R$ in the generator; (3) $G_{ER}$: Both pathways $G_E$ and $G_R$.

From the results, among the single pathways (either $G_E$ or $G_R$), there is no consistent winner and it depends on the dataset. However, it is clear that the use of both pathways in the generator ($G_{ER}$) outperforms using just a single pathway. Thus, this addresses RQ1 and shows that diversifying the negative triplets with the two-way generator can improve model performance and improve the overall informativeness of the negative triplets. 

Furthermore, by comparing $G_{ER}$ (\ie, both pathways without FiLM) and the proposed model \model\ (\ie, both pathways with FiLM), our model obtains a significant lead in performance. This addresses RQ2 and shows the effectiveness of our adaptive design using FiLM.

\stitle{Parameter sensitivity (RQ3).}
To answer RQ3, we perform a parameter sensitivity analysis. We first analyze how the number of negative triplets per positive triplet, $N_s$, can impact model performance. As shown in Figure~\ref{figure2}(b), as we increase $N_s$ on each dataset, we consistently observe that the MRR performance improves and peaks at $N_s=20$. A larger $N_s$ allows for greater diversity, which explains the initial increase in performance. However, when $N_s\ge 20$, performance starts to plateau or even deteriorate, due to imbalanced training data. 

Next, we investigate the impact of adaptive regularization controlled by $\lambda$ in Figure~\ref{figure2}(c). Generally, having such a regularization (\ie, $\lambda > 0$) avoids excessive scaling and shifting from the FiLM layer, and thus reduces overfitting to individual entities or relations. In particular, the MRR performance improves as $\lambda$ increases and achieves the most optimal performance for all three datasets when $\lambda$ is around 1e-4. 
As the optimal values of $N_s$ and $\lambda$ are largely stable across the three datasets, our model is not sensitive to these hyperparameter settings, and potentially requires less effort in hyperparameter tuning.
We also note that the performance on the UMLS datasets tends to be more sensitive to changes in both parameters. This could be because UMLS is a smaller dataset than the other two, containing only 5,216 positive triplets in training and this increases the risk of overfitting to certain settings in general.

\begin{figure}[t]
\captionsetup{justification=centering,singlelinecheck=false}
    \minipage{0.33\textwidth}
      \includegraphics[width=1\textwidth]{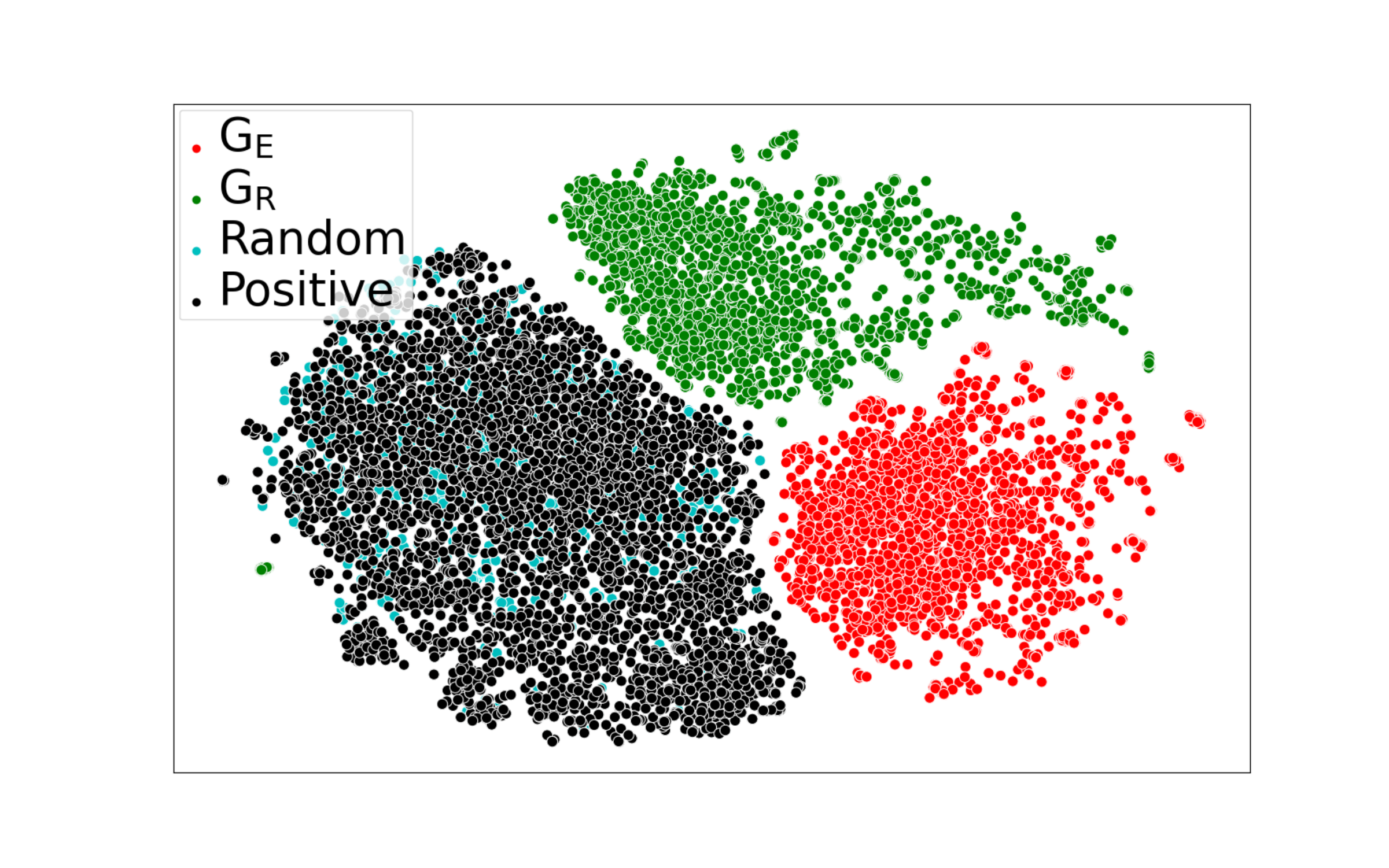}%
      \vspace{-4mm}%
      \caption*{(a) \ent{hasPart}\\(WN18RR)}
    \endminipage%
    \minipage{0.33\textwidth}
      \includegraphics[width=1\textwidth]{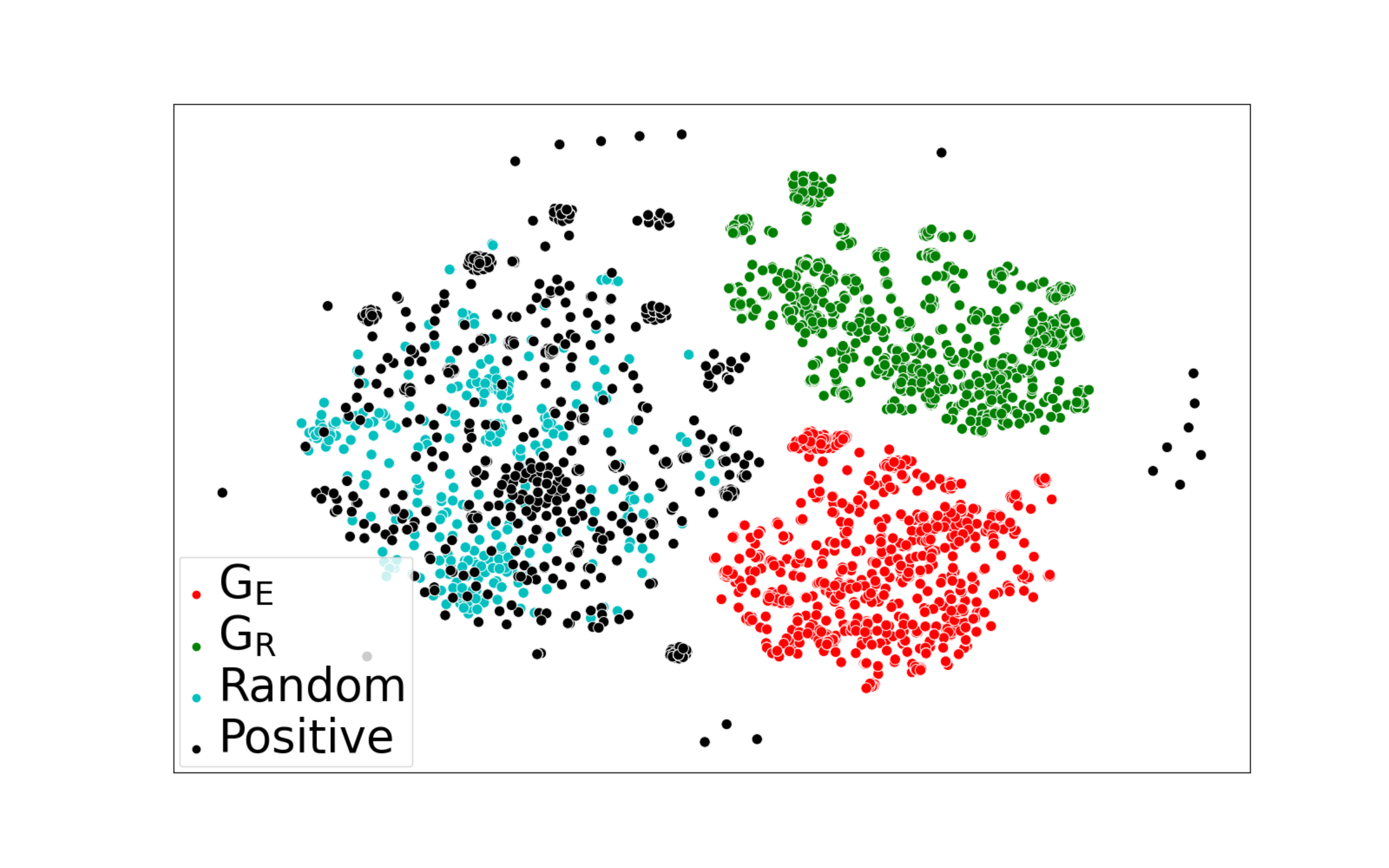}%
      \vspace{-4mm}%
      \caption*{(b) \ent{animalType}\\(NELL-995)}
    \endminipage%
    \minipage{0.33\textwidth}
      \includegraphics[width=1\textwidth]{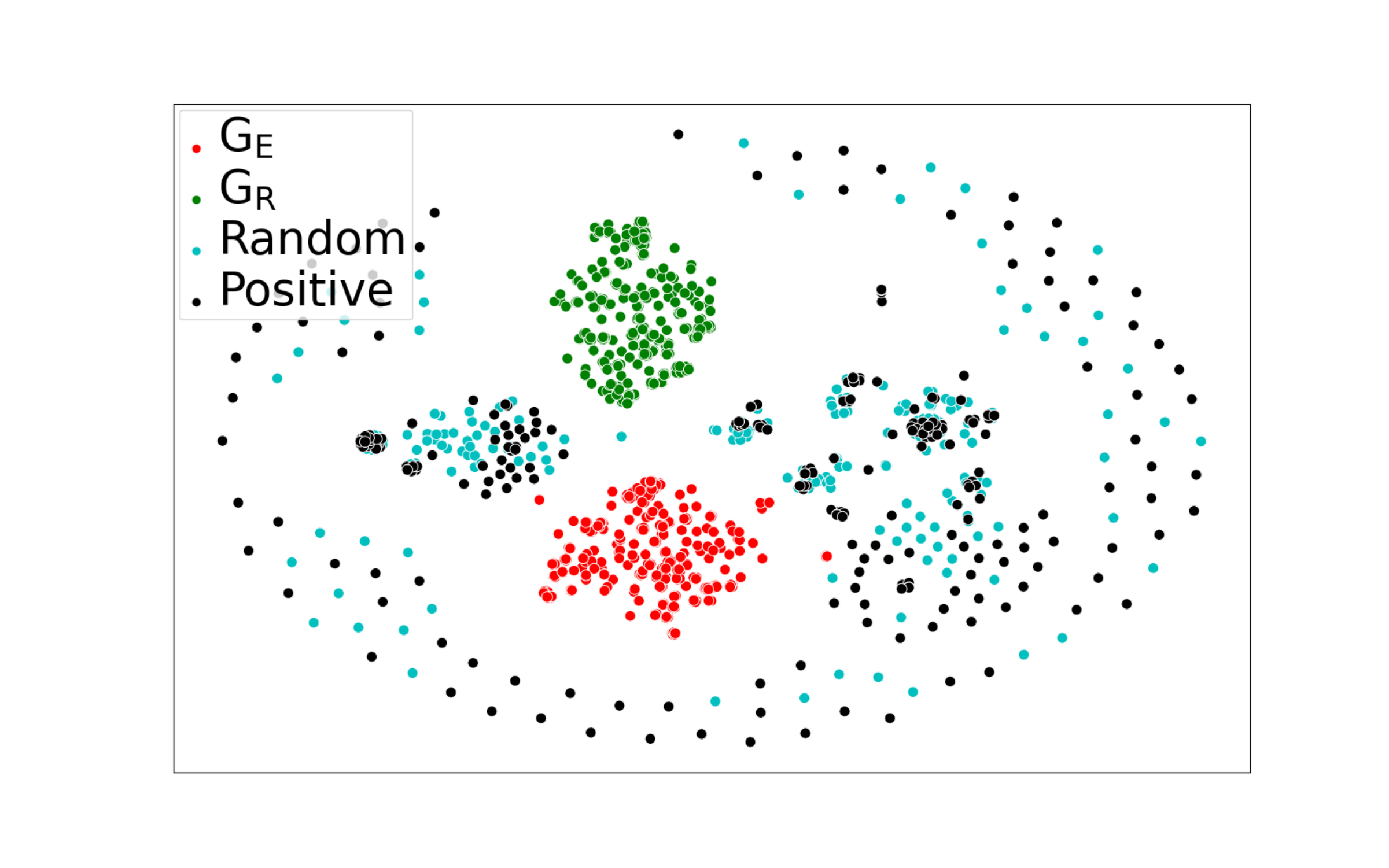}%
      \vspace{-4mm}%
      \caption*{(c) \ent{isA}\\(UMLS)}
    \endminipage
\captionsetup{justification=centering,singlelinecheck=false}
\caption{Visualization of diversity. Best viewed in color.}\label{fig:casestudy1}
\end{figure}

\stitle{Case studies (RQ4).} We conduct a qualitative evaluation of \model\ to demonstrate the diversity of negative triplets generated by \model. 
While the ablation study has demonstrated the importance of diversity through improved model performance with the two-way generator, we further present a few case studies of how \model\ can produce more diverse examples than uniform random sampling (RNS). In Figures~\ref{fig:casestudy1}, we visualize the positive and negative tail entities w.r.t.~a given relation and all its head entities on each dataset. More specifically, each point represents one tail entity, which can be a positive (real) tail entity, or a negative tail entity. The negative entity can be generated by one of the pathways $G_E$ or $G_R$ of the generator, or randomly sampled by RNS. The high-dimensional embedding space is projected onto a Cartesian plane using the $t$-SNE algorithm \cite{van2008visualizing}.
In Figure~\ref{fig:casestudy1}, we compare the diversity of negative entities generated by \model\ with that of RNS-based negative entities. For our case study, we select one relation for each dataset, namely, \ent{hasPart} on WN18RR, \ent{animalType} on NELL-995 and \ent{isA} on UMLS, so that all the positive and negative tail entities for a common relation (and the same original head entities) can be contrasted in one visualization. The results show that \model\ could provide more diverse negative entities for model training, where those generated by $G_E$ and $G_R$ occupy different subspaces from the positive entities. In contrast, RNS lacks diversity and samples negative entities in the same subspace as positive entities. This could even potentially contribute to false negative triplets as they are not well separated from the real ones.

\section{Conclusion}\label{sec:conclusion}
In this work, we introduced $\model$, a negative sampling strategy for knowledge graph embedding that explicitly accounts for the informativeness of negative triplets. On one hand, we proposed a two-way generator to increase the overall informativeness by diversifying the negative triplets based on their association with not only entities but also relations. On the other hand, we adapt the global generator model into local models, which generate negative triplets in a finer-grained manner to improve their individual informativeness. 
Empirically, $\model$ has outperformed state-of-the-art baselines on three benchmark knowledge graphs through both quantitative and qualitative experiments. 

\bibliographystyle{splncs04}
\bibliography{DANS-refs}
\end{document}